# The Generative Energy Arena (GEA): Incorporating Energy Awareness in Large Language Model (LLM) Human Evaluations

Carlos Arriaga[a], Gonzalo Martínez[a], Eneko Sendin[a], Javier Conde[a], Pedro Reviriego[a*]

[a]*ETSI de Telecomunicación, Universidad Politécnica de Madrid, Spain*

*Corresponding Author

**Abstract**

The evaluation of large language models is a complex task, in which several approaches have been proposed. The most common is the use of automated benchmarks in which LLMs have to answer multiple-choice questions of different topics. However, this method has certain limitations, being the most concerning, the poor correlation with the humans. An alternative approach, is to have humans evaluate the LLMs. This poses scalability issues as there is a large and growing number of models to evaluate making it impractical (and costly) to run traditional studies based on recruiting a number of evaluators and having them rank the responses of the models. An alternative approach is the use of public arenas, such as the popular LM arena, on which any user can freely evaluate models on any question and rank the responses of two models. The results are then elaborated into a model ranking. An increasingly important aspect of LLMs is their energy consumption and, therefore, evaluating how energy awareness influences the decisions of humans in selecting a model is of interest. In this paper, we present GEA, the Generative Energy Arena, an arena that incorporates information on the energy consumption of the model in the evaluation process. Preliminary results obtained with GEA are also presented, showing that for most questions, when users are aware of the energy consumption, they favor smaller and more energy efficient models. This suggests that for most user interactions, the extra cost and energy incurred by the more complex and top-performing models do not provide an increase in the perceived quality of the responses that justifies their use.

*Keywords:* LLMs, Evaluation, Energy, Sustainability

# 1. Introduction

Evaluating the performance of Large Language Models (LLMs) is a complex task with many dimensions Chang et al. (2024). For example, multiple choice tests are used to evaluate the ability of LLMs to answer questions on almost any topic we can think of Hendrycks et al. (2020). These tests can be automated and run at scale on many models even if each benchmark has tens of thousands of questions. However, these tests have several limitations, for example, the LLMs may have biases when selecting the options Zheng et al. (2023) or the questions may have been part of the training dataset of the models leading to data contamination Xu et al. (2024). The LLM have also saturated many of the becnhmarks answering correctly most of the questions. Those issues can be partly addressed by not making public the questions and making them harder Phan et al. (2025). However, benchmarks still have limitations, for example they do fully not capture how LLMs generate text and how it aligns with human preferences.

Another approach to evaluate performance is to have an LLM evaluate the responses of other LLMs to different questions Zheng et al. (2024). The use of an LLM as the judge also enables testing at scale but also has limitations. For example, LLMs can introduce biases in evaluation Panickssery et al. (2024) and their judgments may not be fully aligned with humans. To overcome these issues, the obvious solution is to resort to human evaluation. However, human evaluation faces many challenges, LLMs are released almost every week, and conducting human evaluations on thousands of questions requires time and effort. There is also the question of who runs the evaluation and ensures that the results are comparable.

To overcome these issues, a different and innovative approach is to have users evaluate the models freely in open arenas such as the LM arena Chiang et al. (2024). In these arenas, users ask a question and the answers of two models that are not known to the user are provided asking the question to judge if one is better than the other. The results of the votes are used to rank models in the same way as the results of matches are used to rank chess players. The arenas enable a ranking of models based on human preferences; for example, the LM arena is widely used as one of the methods to compare models. The arenas have also limitations as the questions asked to the models are not controlled and the same applies to the profile of the evaluators, and the arenas can potentially be manipulated Min et al. (2025).

Another important aspect of LLMs is their energy consumption, which is very large for training which requires massive processing of data, but also for inference, as models are massively used Morrison et al. (2025). Significant efforts are being made to better understand and optimize the energy dissipation of LLMs Wu et al. (2025),Maliakel et al. (2025).

Interestingly, LLM performance evaluation by humans and LLM energy consumption are intertwined. For example, if users have information on the energy consumption of the models, how would this influence their votes when comparing the models? This is interesting both to understand the preferences of the users and also the performance versus cost trade-offs when developing new LLMs. An initial attempt was made as part of Chung et al. (2025) by implementing a Colosseum to compare LLMs that took into account information on energy consumption[1]. However, it seems to be restricted to open-weight models; at the time of writing this paper, it was not in operation, and more importantly, no analysis of the impact of energy information on the users' preference has been made public. To address those issues and study the impact of energy information on users' preferences, in this paper, an initial step is taken by presenting the Generative Energy Arena (GEA): an LLM arena that provides users with information on the relative energy consumption of LLMs when voting. The results of the testing with users are also discussed showing that energy information has a significant impact on their decisions.

## 2. Generative Energy Arena (GEA): design choices

The design of an arena capable of evaluating the impact of users being aware of the energy consumption of the models when ranking their responses poses a number of issues:

1. The energy consumption of the models is not publicly available for proprietary models and even for open weights models it depends on the hardware platform they are run on.
2. The information on energy consumption has to be given to the user when ranking the responses in such a way that it does not introduce a bias towards selecting lower energy models by default.

---

[1]See https://ml.energy/leaderboard

3. New metrics have to be defined to measure the impact of energy awareness as the goal is not to rank models but to understand the impact of energy awareness.

In the following subsections, each of those issues is discussed in detail, presenting the solution adopted in the proposed GEA.

*2.1. Energy consumption information*

The limited availability of information on the power consumption of proprietary LLMs poses a limitation to make for example comparisons between OpenAI's GPTs and Google's Gemini models. Even making comparisons between pairs of open-weights models can be problematic as their energy consumption varies depending on the hardware platform they are run on and different configuration parameters such as the batch size, quantization or even the verbosity of the models Chung et al. (2025),Conde et al. (2024).

However, an interesting observation is that to assess the impact of energy consumption on user decisions, one possible strategy could be to compare models that are as similar as possible so that the difference in performance is mostly due to model size. For example, two models from the same family but with different sizes are expected to share much of the training data and architectural design so that it is only the scale that changes the performance and the energy consumption. This approach will also ensure that the energy consumption is larger for the larger model facilitating relative comparisons that can be presented in a simple manner to the arena's users. Simplicity is also an important design goal as most users will not be experts on LLMs or energy metrics.

The scheme implemented in GEA is to provide information on energy consumption in relative terms, that is when two models are evaluated the arena will provide information on which of the two used more energy without providing absolute values. This gives users the information required to understand the energy implications of using each model without having to go into details which are not available for most models. To enable those relative comparison making the rest of the factors as similar as possible only models of different sizes from the same family are compared. For example, for GPT4.1 the user can get the responses of GPT4.1, GPT4.1-mini and GPT4.1-nano with the information that the energy consumption of GPT4.1 is larger than that of GPT4.1-mini which in turn is larger than that of GPT4.1-nano.

## 2.2. Presenting the energy information to the users

Another important element in the design of the arena is how to present the energy information to the users. If the information is given beforehand it can bias the users towards selecting the models with the lowest energy consumption before evaluating the quality of the LLM responses. To try to avoid this issue and similarly to other works Chung et al. (2025), in the GEA users go through a two-step evaluation process:

1. They ask a question to two unknown models and are asked to select the best answer.
2. If the user selected as best answer that from the model with the largest energy consumption, then the user is asked "Knowing that the other response consumes less energy, would you change your choice assuming a loss in quality?"

This strategy ensures that users first evaluate the responses based only on their quality and then the impact of energy awareness in an independent step which not only avoids potential biases in the first decision but also facilitates measuring the impact of the energy information as we have the user choices without and with that information.

## 2.3. Impact metrics

To analyze the impact of energy awareness on the decisions we have to define metrics that capture that impact. As our comparisons are based on pairs of models of a given family, we will use the fraction of the responses for which the users changed their vote after having the energy information which we will denote as $E_c$.

In addition, we will also compare the win rates of the two models before and after having the energy information to get an indication of the overall impact of energy awareness on the relative model performance. In more detail, given a pair of models that we denote as large $L$ and small $S$ with initial win rates $W_L$, $W_S$ and tie rate $T$ the winning rates after having into account energy efficiency will be computed as:

$$W_S(E) = W_S + T + W_L \cdot E_c \quad (1)$$

$$W_L(E) = W_L \cdot (1 - E_c) \quad (2)$$

## 3. Generative Energy Arena (GEA): implementation

The arena has been implemented as a space in Hugging Face and currently supports comparisons between models of the following families:

- GPT-4o-2024-08-06 and GPT-4o-mini-2024-07-18.
- GPT-4.1-2025-04-14 and GPT-4.1-mini-2025-04-14.
- Claude Sonnet 3.5 and Haiku 3.5.
- Llama3-70b-versatile and Llama3-8b-8192.

When a user accesses the arena, a family of models is chosen randomly and if there are more than two models in the family, a pair is selected also randomly. The user responses are stored in a database that can be accessed for data processing and analysis, and the code for the arena is publicly available. The user interface of the GEA is shown in Figure 1. The landing page provides information on the project and how the arena works, and provides access to the arena itself on a second page. The results are also shown on a separate page per model family showing the percentage of votes before and after the energy consumption information was given to the user.

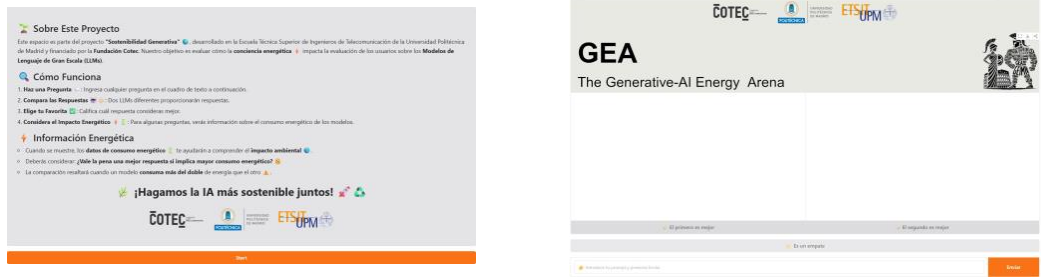

Figure 1: User Interface of the Generative Energy Arena (GEA): Landing page (left) and Main page (right)

## 4. Generative Energy Arena (GEA): results and analysis

The arena has been used as part of a Massive Open Online Course offered by the Universidad Politécnica de Madrid. In this course, students are asked to evaluate five questions provided in an assignment and another five created by them on the GEA. This enables us to estimate the fraction of responses that correspond to students of the MOOC who should be knowledgeable about LLMs. For the rest of the responses we do not have any information on the users as the arena is open and anyone can access it.

At the time of writing this paper a total of 694 questions have been evaluated in the GEA of which 295 correspond to the questions provided in the MOOC assignment. These questions have been formulated in Spanish and are listed below with the English translation:

1. "Invéntate un eslogan para promocionar un XXXX" (cambiar XXXX por un producto).
   "Invent a slogan to promote a XXXX" (replace XXXX with a product).
2. "Explícame qué es el parámetro Top-p en los LLM".
   "Explain to me what the Top-p parameter is in an LLM."
3. "Escribe un poema de 4 versos en el que juntando la primera letra de cada verso forma una palabra".
   "Write a 4-line poem in which the first letter of each line forms a word."
4. "Dime lo que sabes sobre el pueblo XXX" (cambiar XXX por el nombre de un pueblo).
   Tell me what you know about the town XXX" (replace XXX with the name of a town).
5. "Dame una receta que pueda preparar con estos ingredientes: XX, XX, XX..." (cambiar XX por ingredientes).
   "Give me a recipe I can make with these ingredients: XX, XX, XX..." (replace XX with ingredients).

These questions are a significant fraction of the sample analysis that accounts for 42.5% of the total. Since the students have to ask five more questions written by them, we estimate that at least 83% of the questions were made by the MOOC students.

The fraction of responses that changed after having the energy information is shown in Figure 2 for the different families of models. It can be seen that there is a significant change in votes when users are aware of the relative energy consumption of the model, which ranges from 41% to 52%, averaging around 46%. This shows that energy awareness should be incorporated in human evaluations, as it can be an important factor in the decision-making process.

The winning rates of the two models for the different pairs before and after having the energy information are shown in Figure 3. When energy is taken into account, it can be seen that smaller models have higher winning rates over their larger counterparts. In more detail, the initial preference between models is comparable with less than a 2% difference. In contrast, the winning rates after accounting for the back-down rate significantly favor the smaller models (blue bars), since the users choose them more than 75% of the time. This is an interesting result: the use of larger LLMs is not worth it in most cases from the user's perspective.

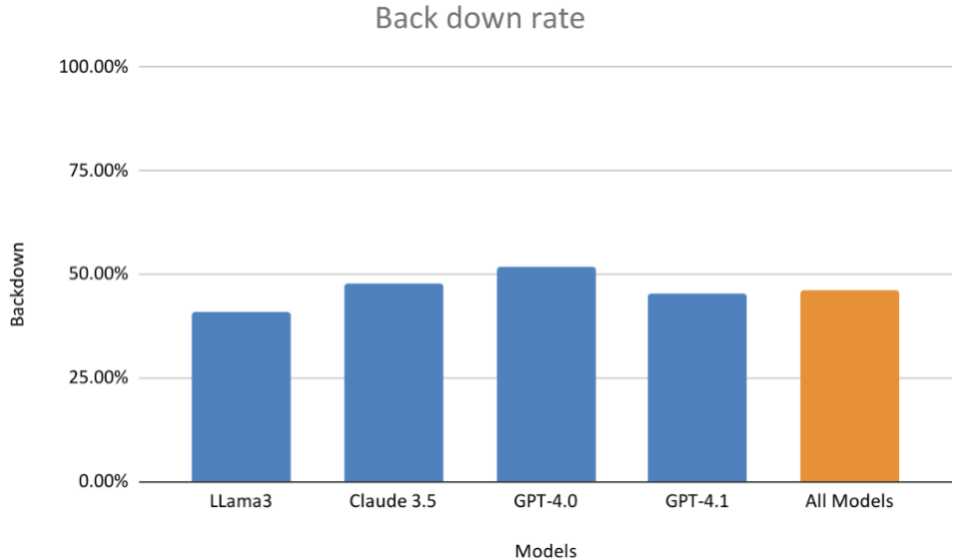

Figure 2: Fraction of responses changed after having the energy information ($E_c$)

In Figure 4 we can observe the winning rates of the pairs of models separated into the four different families of models. The difference between the pairs of models is highlighted in the initial choice of models: while the large models of Llama3 are chosen more often, the preference between models in the Claude family is not clear, and in the GPT families the smaller models are preferred by the user. It is important to note that even before disclosing the energy employed to answer, in some model families there is an initial and pronounced preference for the small model. This can be due to the type of question asked, or to the profile of the GEA users, mostly MOOC students.

This analysis shows how the use of larger and more energy hungry LLMs is only worth it for specific questions while for the majority of users' interactions, in our case, a smaller model is preferred by the users when they are aware of the LLMs energy consumption.

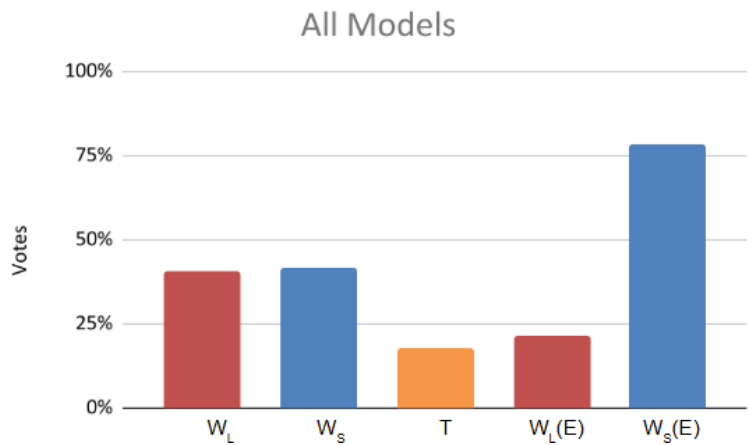

Figure 3: Average winning rates across all the models before and after having the energy information ($W_L$, $W_S$ and $W_L(E)$, $W_S(E)$)

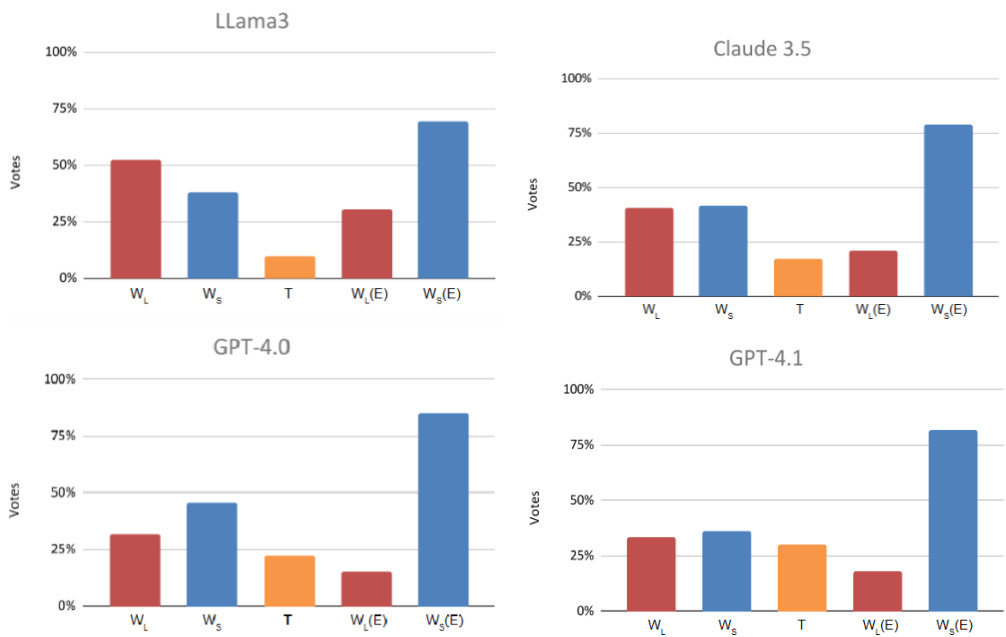

Figure 4: Winning rates the models before and after having the energy information ($W_L$, $W_S$ and $W_L(E)$, $W_S(E)$)

## 5. Limitations

The results and analysis presented in this work are just an initial step towards understanding how energy consumption awareness influences user decisions when selecting an LLM. In particular, this work has several limitations that may limit the general applicability of our conclusions:

1. The number of questions and users is small, only a few hundred; ideally, thousands of questions should be used for analysis.
2. The number of LLMs evaluated is also limited to models from three major companies; ideally, a larger range of LLMs should be evaluated.
3. The analysis is done in only one language, Spanish; ideally different languages should be tested.
4. The analysis is done for all question types; an independent analysis for each type of question may provide additional information. For example, the additional energy cost of larger models may be perceived by users as useful for certain categories or types of questions.

## 6. Conclusion

This paper has analyzed the impact of energy awareness in LLM human evaluation. The results show that having information on the relative energy consumption of the LLMs has a significant impact on user choices. In fact, the winning rates of the smaller models are higher than those of the larger models when both are compared. This suggests that LLMs have reached a point in which smaller versions of the models are sufficient to answer many users' questions, and more advanced and larger models only add value for specific questions and queries. This has implications for the development and deployment of LLMs and also for LLM evaluation. In particular, it seems that energy information is a critical factor for human evaluation and should be included in future evaluation methodologies. Further work is needed to better understand the impact of energy awareness in user decisions. This includes collecting and analyzing a larger sample of questions and also having different user profiles and types of questions. The evaluation should also be extended to a larger set of LLMs covering the entire performance range, and if possible, done in several languages.

# Acknowledgment

This work was supported by the FUN4DATE (PID2022-136684OB-C22) and SMARTY (PCI2024-153434) projects funded by the Spanish Agencia Estatal de Investigación (AEI) 10.13039/501100011033, by the Chips Act Joint Undertaking project SMARTY (Grant no. 101140087) and by the Cotec Foundation.